**Seeing the Unseen: Errors and Bias in Visual Datasets**

Hongrui (Kevin) Jin
kevin.jin@student.uva.nl

# 1. Introduction

From face recognition in smartphones to automatic routing on self-driving cars, machine vision algorithms lie in the core of these features. These systems solve image based tasks by identifying and understanding objects, subsequently making decisions from these information. A large set of images where the featured objects were labelled, known as datasets, are commonly used to develop and enhance machine vision algorithms (Cox 2016). However, errors in datasets are usually induced or even magnified in algorithms, at times resulting in issues such as recognising black people as gorillas and misrepresenting ethnicities in search results (Nieva 2015; Prabhu and Birhane 2020). This essay tracks the errors in datasets and their impacts, revealing that a flawed dataset could be a result of limited categories, incomprehensive sourcing and poor classification.

## 2. Relevant Concepts

### 2.1 Machine Learning Algorithm and Datasets

**Machine learning algorithms** enables devices to detect, identify or produce objects from an image or visual system (Paglen 2014). These algorithms are built based on a predetermined **dataset** that specifies what and where an object is present in a given image. In other words, a relationship between the input image and output categories must be provided externally to develop a machine vision system. This highlights the supervised learning roots of machine vision algorithms (Russell and Norvig 2021). The specification process in datasets, or "tagging", abstracts data by extracting information from images, and transforming them to texts (Pasquinelli 2022). When creating image datasets, developers have to decide where to source images, which classification system to use, and how to apply this classification to millions of images (Prabhu and Birhane 2020). These decisions affect the quality of datasets, thereby influencing algorithms trained by them.

### 2.2 Perception and Bias

In his book The Logic of Perception (1983), Irvin Rock described **perception** as an "intelligent thought-like process". This assumption is further completed by Brian Rogers, referring to perception as an indirect process that is based on insufficient sensory information, and meanings are added rather than intrinsic (Rogers 2017). Datasets translates images to textual information, and could be done manually, or automatically, utilising previous algorithms. It is also possible to use both methods combined to balance efficiency and accuracy. This process relies on and imitates human perception to add meaning to images, as a result inheriting human perception errors in datasets. **Errors and bias**, however, can also be introduced depending on the sources of images, such as selection bias induced by only using images of a certain website, or capture bias due to the intention of photographers (Torralba and Efros 2011). Bias and errors in datasets are magnified by the enormous amount of images included, leading to inaccurate machine vision algorithms.

## 3. Methodology

A total of five open access visual datasets with different features are investigated to reveal their induced bias and errors. A varied range of datasets consisting of crowdsourced collections (ImageMonkey), labeled videos (YouTube-8M), faces (MS-CELEB-1M), and general purpose datasets (Open Images, ImageNet) are intentionally selected to achieve a comprehensive analysis. Their technical publications are also examined if available. These documentations are usually produced by the developers of datasets, making them an authentic perspective of how visual materials were collected and classified. By focusing on selection and classification techniques in aforementioned databases with regards to their contexts, bias and errors could be identified, compared and evaluated. Additionally, any effort to reduce bias in selected datasets is also reviewed as an attempt to inspire future data refinements.

While open access image datasets might contain less images and annotations comparing to their commercial alternatives, they are more suitable for investigation purposes thanks to their transparency and accessibility. Commercial datasets such as the JFT-300M and JFT-3B by Google, on the other hand, are mostly proprietary and their development methods unknown to the public, making it challenging to address their bias and errors effectively.

## 4. Analysis and Evaluation

### 4.1 To Describe Everything: Catalogue Systems

Visual datasets translates visual data to textual information based on a given catalogue, or classification system (Malevé 2019). The ability of a dataset to describe a visual artefact is therefore dependent on the vocabulary available in the classification system it employs. A bigger, general catalogue might resolve visual data extensively, yet it significantly challenges the later classification and labelling process. Large classification system is found in ImageNet, which claimed to already cover 5247 categories in 2009 (Deng et al. 2009), as well as YouTube-8M that annotates 4800 classes (Abu-El-Haija et al. 2016). While these datasets could more accurately describe visual data, it is not always possible to provide sufficient visual data for all categories. YouTube-8M, for example, documented almost 3 million occurrence of "Art and Entertainment" in its dataset, while only having around 10,000 videos labelled with "Finance" (Abu-El-Haija et al. 2016). Training on this unbalanced representation would likely result in an algorithm that excels at identifying entertainment videos but fails to recognise underrepresented contents such as finance. In ImageNet, the dataset reportedly had around 50% categories with more than 500 images and over 20% of the categories with "very few images", revealing a similar disproportional representation. In a more detailed investigation report on ImageNet by Kate Crawford and Trevor Paglen (2019), categorisation were found to be even more problematic: the dataset adapted noun categories from WordNet, a comprehensive word catalogue developed in the 1980s. However, this classification system contained outdated socio-cultural assumptions that invited a prejudiced categorisation, such as only recognising "male" and "female" as natural, and bisexual was found under the "sensualist" hierarchy. This elaborate yet biased source for classification could lead to unjustified image recognition results, exemplified by the *ImageNet Roulette* algorithm that outputs classifiers from faces. The algorithm describes faces regardless their context, sometimes with disturbing attributes such as "gook" and "criminal" (Wong 2019; Artforum 2019).

A smaller classification system is easier to work with, but may over abstract information from images, resulting in ambiguity and difficult decisions such as intentionally omitting certain category. MS-CELEB-1M utilises a categorisation system that only features celebrities enlisted on the knowledge base "Freebase" and ranked them based their appearance on the web (Guo et al. 2016a). However, this knowledge base in only available in English, thereby limiting the inclusion of individuals with non-Latin names, for their web appearance might be in another language. Additionally, filtering "celebrities" based on their number of occurrence on the Internet is hardly accurate, for individuals renown before the age of Internet, or in countries with limited Internet access would be significantly underrepresented. This limited context resulted in a database that is "overwhelmingly white and male" (Harvey 2021) that fails to consider individuals of other ethnicities and backgrounds. Classification systems determines the upcoming process of collecting data, as they provide the keywords for automated search inquiries, and at large determine what to include in datasets (see section **3.2**).

**4.2 Tunnel Vision: Sourcing Visual Data**

Once a classification system is determined, visual data needs to be gathered to substantiate each category. Constructing datasets requires an enormous amount of visual media artefacts. To attain data in such quantity efficiently, data can hardly be produced on demand; automated retrieval are used to scrape images and videos from the Internet. Examining the reported data extraction measures of the 5 selected datasets, it appears visual data are usually collected from various search engines and hosting platforms. Both ImageNet and MS-CELEB-1M reported obtaining images from search engine. However, exact search methods vary. MS-CELEB executed all search enquiries in English (Guo et al. 2016b), while ImageNet utilised multiple search engines and inquired in four languages (Li 2010). This difference is also a result of varying classification systems, since Freebase, which empowered MS-CELEB, only have English entries (Bollacker et al. 2008).

Searching in only English increases language bias by returning English web results that fails to represent visual artefacts in non-English worlds (Spencer and Heneghan 2017). Open Images and YouTube-8M extracts data from media hosting sites such as Flickr and Youtube (Abu-El-Haija et al. 2016; Krasin et al. 2017). These hosting platforms were purposefully used for their higher visual quality, so high resolution media could be collected to rigorously train algorithms. Nevertheless, acquisitions from specialised platforms induce **capture bias**, a particular form of selection bias that affects individuals creating media artefacts (Torralba and Efros 2011). This is best exemplified by an imaginary photographer taking photos: their photos filters visual elements and perspectives are likely guided and limited by aesthetic intention, so images of "people" are always captured *around* the subject, instead of from above and below. As for YouTube, selection bias occurs since only videos with "at least 1,000" views between 120 to 500 secs are collected (Abu-El-Haija et al.). Such omission admittedly allows for a comprehensive dataset under a reasonable size, yet view counts and time lengths do not always determine the "quality" of media artefacts, especially in the case of YouTube-8M with more than 4800 classes where some categories had significantly less videos than others. Regarding the four datasets above, they are affected by what I refer to as *tunnel vision*, a term originally used to describe visual distortion (see **Fig. 1**): utilising a restricted search technique on one hand signifies certain objects and features in the dataset, yet hinders a comprehensive visual portrayal of the classification system.

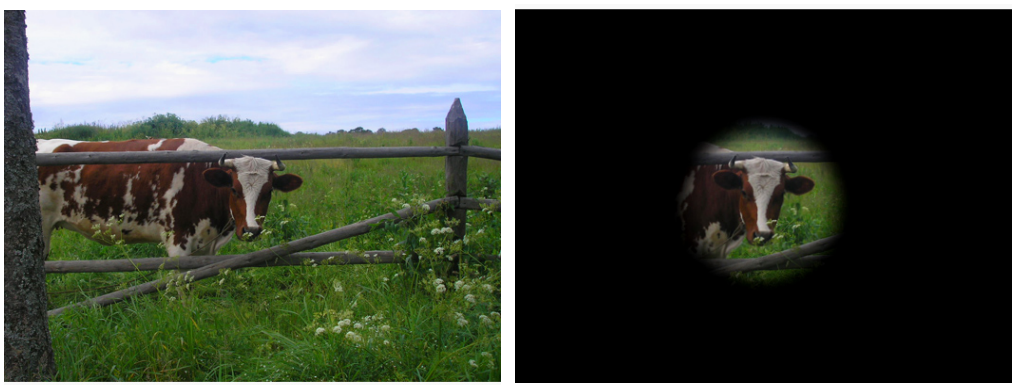

**Figure 1** An example of tunnel vision (Scampetsky 2009)

ImageMonkey proposed another method for acquiring visual data. Unlike aforementioned datasets, ImageMonkey does not collect data from other platforms or search engines; the dataset is entirely crowdsourced and gathers images solely from donations. This collection technique opens access to everyone rather than specifying platforms, thereby eliminating the bias induced in the selecting systems of other datasets such as the YouTube recommendation algorithm. Users worldwide can submit their photos and index them, allowing for a balanced socio-cultural representation across countries in comparison to English exclusive datasets such as MS-CELEB. However, manual uploads and passive donations render rapid development unfeasible: ImageMonkey only accumulated around 10K images in five years, where automated retrieval could extract millions of images in a second. This size limitation in turn affects the accuracy of algorithms trained on it (Malevé 2019), making ImageMonkey less usable than others.

**4.3 A Million Times: Labelling and Classification**

Labelling matches visual data to categories; the accuracy this process contributes to the accuracy of a trained algorithm. Apart from manually labelling data using human perception, it is common to use previous machine vision algorithms to assist and speed this process. Developers of ImageNet utilised Amazon Turk Workers (AMTs) to "tag" images (Prabhu and Birhane 2020), and ImageMonkey encourages users to label images via gamification (Bernhard n.d.). While manual labelling benefits from human perception and knowledge, they are criticised to induce human errors and socio-cultural influences. Carl Vondrick et al. (2014) examined the bias introduced by AMTs in the case of identifying sports balls: Americans imagines orange and brown balls, comparing to Indians imaging red ones (see **Fig. 2**). This was explained by the popular sports in different countries, such as cricket in India, or football and baseball in United States.

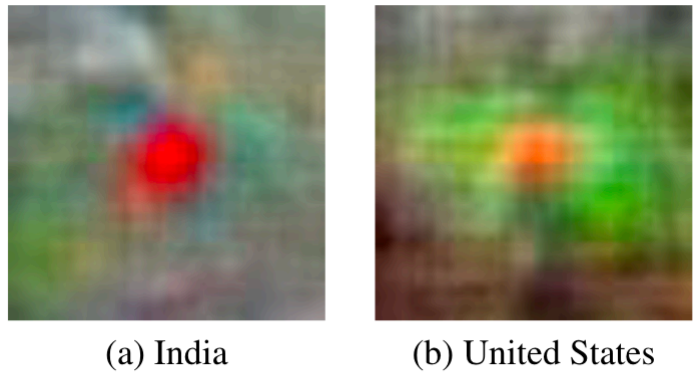

(a) India  (b) United States

**Figure 2** Colours of imagined balls, visualised by Indian and American AMTs. (Vondrick et al. 2014)

In addition to cultural influences, manual labelling induces errors by exhausting human attention. Sen He et al. (2019) measured eye movements of individuals during free-viewing and timed labelling tasks, discovering that timed observation resulted in a disperse of focus and a loss of details in captions produced. It is reasonable to assume such errors also occur when using AMTs, as these tasks are repetitive and timed. Other datasets such as MS-CELEB and Open Images employed image recognition algorithms to label images and verify datasets. While this procedure enables faster development, their accuracy could not be evaluated due to the sheer amount of data present and the dependence on previous datasets. Northcutt, Athalye, and Mueller (2021) reported an error rate of at least 3.3% across 10 datasets, with ImageNet owning at least 6% mislabelled images. Vijay Vasudevan et al. (2022) discovered that over 60% of the image annotation ImageNet deemed incorrect were false positives. More importantly, the errors from one dataset could be carried on to another if datasets use flawed existing ones for benchmark (Crawford 2021). These reports confirm the alarming amount of errors introduced from manual tagging since AMTs are the only source of labelling in ImageNet.

## 5. Conclusion

Errors and bias in the making of datasets are discovered and evaluated through the analysis of 5 open visual datasets of various features and sizes. I suggest that errors and bias could occur in 3 major stages of dataset production: 1) catalogue system could induce strong socio-cultural prejudice and cause a disproportional representation of data; 2) limiting and ranking source data assume bias from existing platforms; 3) manual and algorithmic labelling generates human errors and false positives during benchmarks. Concerning machine vision algorithms and datasets "learn" from existing ones, these findings might motivate developers and researchers to refine current methods to curate unprejudiced inclusive datasets for the ever-growing technologies of machine vision.

# Bibliography

Abu-El-Haija, Sami, Nisarg Kothari, Joonseok Lee, Paul Natsev, George Toderici, Balakrishnan Varadarajan, and Sudheendra Vijayanarasimhan. 2016. 'YouTube-8M: A Large-Scale Video Classification Benchmark'. arXiv:1609.08675. arXiv. https://doi.org/10.48550/arXiv.1609.08675.

Artforum. 2019. 'Image Database Purges 600K Photos After Trevor Paglen Project Reveals Biases'. *Artforum* (blog). 24 September 2019. https://www.artforum.com/news/trevor-paglen-and-kate-crawford-s-reveals-its-racist-misogynistic-biases-imagenet-to-purge-600-000-facial-images-from-database-80829.

Bernhard, B. n.d. 'ImageMonkey'. Accessed 2 June 2022. https://imagemonkey.io/.

Bollacker, Kurt, Colin Evans, Praveen Paritosh, Tim Sturge, and Jamie Taylor. 2008. 'Freebase: A Collaboratively Created Graph Database for Structuring Human Knowledge'. In *Proceedings of the 2008 ACM SIGMOD International Conference on Management of Data*, 1247–50. SIGMOD '08. New York, NY, USA: Association for Computing Machinery. https://doi.org/10.1145/1376616.1376746.

Cox, Geoff. 2016. 'Ways of Machine Seeing'. November 2016. https://unthinking.photography/articles/ways-of-machine-seeing.

Crawford, Kate. 2021. *Atlas of AI: Power, Politics, and the Planetary Costs of Artificial Intelligence*. New Haven: Yale University Press.

Deng, Jia, Wei Dong, Richard Socher, Li-Jia Li, Kai Li, and Li Fei-Fei. 2009. 'ImageNet: A Large-Scale Hierarchical Image Database'. In *2009 IEEE Conference on Computer Vision and Pattern Recognition*, 248–55. Miami, FL: IEEE. https://doi.org/10.1109/CVPR.2009.5206848.

Guo, Yandong, Lei Zhang, Yuxiao Hu, Xiaodong He, and Jianfeng Gao. 2016a. 'MS-Celeb-1M: Challenge of Recognizing One Million Celebrities in the Real World'. In . https://www.researchgate.net/publication/295074418_MS-Celeb-1M_Challenge_of_Recognizing_One_Million_Celebrities_in_the_Real_World.
———. 2016b. 'MS-Celeb-1M: A Dataset and Benchmark for Large-Scale Face Recognition'. In . https://www.microsoft.com/en-us/research/publication/ms-celeb-1m-dataset-benchmark-large-scale-face-recognition-2/.

Harvey, Adam. 2021. 'Exposing.Ai: MS-Celeb-1M'. Exposing.Ai. 1 January 2021. https://exposing.ai/datasets/msceleb/.

He, Sen, Hamed Rezazadegan Tavakoli, Ali Borji, and Nicolas Pugeault. 2019. 'Human Attention in Image Captioning: Dataset and Analysis'. In *2019 IEEE/CVF International Conference on Computer Vision (ICCV)*, 8528–37. Seoul, Korea (South): IEEE. https://doi.org/10.1109/ICCV.2019.00862.

Krasin, Ivan, Tom Duerig, Neil Alldrin, Vittorio Ferrari, Sami Abu-El-Haija, Alina Kuznetsova, Hassan Rom, et al. 2017. 'OpenImages: A Public Dataset for Large-Scale Multi-Label and Multi-Class Image Classification.' *Dataset Available from Https://Storage.Googleapis.Com/Openimages/Web/Index.Html*.

Li, Fei-Fei. 2010. 'ImageNet: Crowdsourcing, Benchmarking & Other Cool Things'. Stanford University.

Malevé, Nicolas. 2019. 'An Introduction to Image Datasets'. November 2019. https://unthinking.photography/articles/an-introduction-to-image-datasets.

Nieva, Richard. 2015. 'Google Apologizes for Algorithm Mistakenly Calling Black People "Gorillas"'. CNET. 1 July 2015. https://www.cnet.com/tech/services-and-software/google-apologizes-for-algorithm-mistakenly-calling-black-people-gorillas/.

Northcutt, Curtis G., Anish Athalye, and Jonas Mueller. 2021. 'Pervasive Label Errors in Test Sets Destabilize Machine Learning Benchmarks'. arXiv:2103.14749. arXiv. https://doi.org/10.48550/arXiv.2103.14749.

Paglen, Trevor. 2014. 'Seeing Machines'. Fotomuseum Winterthur. 13 March 2014. https://www.fotomuseum.ch/de/2014/03/13/seeing-machines/.

Pasquinelli, Matteo. 2022. 'From Algorism to Algorithm: A Brief History of Calculation from the Middle Ages to the Present Day'. *Electra*, The Numbers, , February. https://matteopasquinelli.com/from-algorism-to-algorithm-a-brief-history-of-calculation-from-the-middle-ages-to-the-present-day/.

Prabhu, Vinay Uday, and Abeba Birhane. 2020. 'Large Image Datasets: A Pyrrhic Win for Computer Vision?' arXiv:2006.16923. arXiv. https://doi.org/10.48550/arXiv.2006.16923.

Rock, Irvin. 1983. *The Logic of Perception*. MIT Press.

Rogers, Brian. 2017. *Perception: A Very Short Introduction*. https://www.overdrive.com/search?q=BA689E26-B07F-4DBE-8273-D0A8714FC6FD.

Russell, Stuart J., and Peter Norvig. 2021. *Artificial Intelligence: A Modern Approach*. Fourth edition. Pearson Series in Artificial Intelligence. Hoboken: Pearson.

Spencer, EA, and C Heneghan. 2017. 'Language Bias'. In *Catalog of Bias*. https://catalogofbias.org/biases/language-bias/.

Torralba, Antonio, and Alexei A. Efros. 2011. 'Unbiased Look at Dataset Bias'. In *CVPR 2011*, 1521–28. https://doi.org/10.1109/CVPR.2011.5995347.

Vasudevan, Vijay, Benjamin Caine, Raphael Gontijo-Lopes, Sara Fridovich-Keil, and Rebecca Roelofs. 2022. *When Does Dough Become a Bagel? Analyzing the Remaining Mistakes on ImageNet*.

Vondrick, Carl, Hamed Pirsiavash, Aude Oliva, and Antonio Torralba. 2014. 'Learning Visual Biases from Human Imagination', October. https://doi.org/10.48550/arXiv.1410.4627.

Wong, Julia Carrie. 2019. 'The Viral Selfie App ImageNet Roulette Seemed Fun – until It Called Me a Racist Slur'. *The Guardian*, 18 September 2019, sec. Technology. https://www.theguardian.com/technology/2019/sep/17/imagenet-roulette-asian-racist-slur-selfie.